\documentclass[10pt,journal,compsoc]{IEEEtran}

\ifCLASSOPTIONcompsoc
  \usepackage[nocompress]{cite}
\else
  \usepackage{cite}
\fi

\usepackage{ragged2e}

\ifCLASSINFOpdf
  \usepackage[pdftex]{graphicx}
\else
  \usepackage[dvips]{graphicx}
\fi

\usepackage{algorithm}
\usepackage{algorithmic}
\usepackage{amsmath}
\usepackage{multirow}
\usepackage{hyperref}

\begin{document}
%
\title{AugGPT: Leveraging ChatGPT for Text Data Augmentation}

\author{Haixing~Dai{$\ast$},
        Zhengliang~Liu{$\ast$},
        Wenxiong~Liao{$\ast$},
        Xiaoke~Huang,
        Yihan~Cao,
        Zihao~Wu,
        Lin~Zhao,
        Shaochen~Xu,
        Wei~Liu,
        Ninghao~Liu,
        Sheng~Li,
        Dajiang~Zhu,
        Hongmin~Cai, 
        Lichao~Sun,
        Quanzheng~Li,
        Dinggang~Shen,
        Tianming~Liu,
        and~Xiang~Li 

\IEEEcompsocitemizethanks{
\IEEEcompsocthanksitem $\ast$ These authors contributed equally to this paper.
\IEEEcompsocthanksitem Haixing Dai, Zhengliang Liu, Zihao Wu, Lin Zhao, Shaochen Xu, Ninghao Liu, and Tianming Liu are with the School of Computing, University of Georgia, Athens, GA, USA. (e-mail: \{hd54134, zl18864 ,zw63397, lin.zhao, shaochen.xu25, ninghao.liu, tliu\}@uga.edu).
\IEEEcompsocthanksitem Wenxiong Liao, Xiaoke Huang, Hongmin Cai are with the School of Computer Science and Engineering, South China University of Technology, China. (e-mail: \{cswxliao@mail.scut.edu.cn, csxkhuang@mail.scut.edu.cn, hmcai@scut.edu.cn).
\IEEEcompsocthanksitem Yihan Cao and Lichao Sun are with the Department of Computer Science and Engineering, Lehigh University, Bethlehem, PA, USA. (e-mail: \{yihanc@andrew.cmu.edu, lis221@lehigh.edu).
\IEEEcompsocthanksitem Yihan Cao is also with the Heinz College of Information Systems and Public Policy, Carnegie Mellon University, Pittsburgh, PA, USA (e-mail: \{email yihanc@andrew.cmu.edu).
\IEEEcompsocthanksitem Wei Liu is with the Department of Radiation Oncology, Mayo Clinic, Phoenix, AZ, USA. (e-mail: liu.wei@mayo.edu)
\IEEEcompsocthanksitem Sheng Li is with the School of Data Science, University of Virginia, Charlottesville, VA, USA. (email: shengli@virginia.edu) 
\IEEEcompsocthanksitem Dajiang Zhu is with the Department of Computer Science and Engineering, The University of Texas at Arlington, Arlington, TX, USA. (e-mail: dajiang.zhu@uta.edu)
\IEEEcompsocthanksitem Quanzheng Li and Xiang Li are with the Department of Radiology, Massachusetts General Hospital and Harvard Medical School, Boston, MA, USA. (e-mail: li.quanzheng@mgh.harvard.edu, xiangli.shaun@gmail.com)
\IEEEcompsocthanksitem Dinggang Shen is with the School of Biomedical Engineering, ShanghaiTech University, Shanghai 201210, China. He is also with Shanghai United Imaging Intelligence Co., Ltd., Shanghai 200230, China, and Shanghai Clinical Research and Trial Center, Shanghai, 201210, China. (e-mail: Dinggang.Shen@gmail.com)

}
} 

\IEEEtitleabstractindextext{%
\begin{abstract}

\justifying
Text data augmentation is an effective strategy for overcoming the challenge of limited sample sizes in many natural language processing (NLP) tasks. This challenge is especially prominent in the few-shot learning scenario, where the data in the target domain is generally much scarcer and of lowered quality. A natural and widely-used strategy to mitigate such challenges is to perform data augmentation to better capture the data invariance and increase the sample size. However, current text data augmentation methods either can’t ensure the correct labeling of the generated data (lacking faithfulness) or can’t ensure sufficient diversity in the generated data (lacking compactness), or both. Inspired by the recent success of large language models, especially the development of ChatGPT, which demonstrated improved language comprehension abilities, in this work, we propose a text data augmentation approach based on ChatGPT (named AugGPT). AugGPT rephrases each sentence in the training samples into multiple conceptually similar but semantically different samples. The augmented samples can then be used in downstream model training. Experiment results on few-shot learning text classification tasks show the superior performance of the proposed AugGPT approach over state-of-the-art text data augmentation methods in terms of testing accuracy and distribution of the augmented samples.

\end{abstract}

}

\maketitle

\IEEEdisplaynontitleabstractindextext

%
\IEEEpeerreviewmaketitle

\IEEEraisesectionheading{\section{Introduction}\label{sec:introduction}}

\IEEEPARstart{T}{he} effectiveness of natural language processing (NLP) heavily relies on the quality and quantity of the training data. With limited training data available, which is a common issue in practice due to privacy concerns or the cost of annotations, it can be challenging to train an accurate NLP model that generalizes well to unseen samples. The challenge of training data insufficiency is especially prominent in few-shot learning (FSL) scenarios, where the model trained on the original (source) domain data is expected to generalize from only a few examples in the new (target) domain \cite{wang2020generalizing}. Many FSL methods have shown promising results in overcoming this challenge in various tasks. 
Existing FSL methods mainly focus on improving the learning and generalization capability of the model via better architectural design ~\cite{yin2020meta, wang2021transprompt}, leveraging pre-trained language models as the basis and then fine-tuning it using limited samples~\cite{devlin2018bert} with meta-learning~\cite{yin2020meta, lee2022meta} or prompt-based methods~\cite{brown2020language, lester2021power, han2022ptr, wang2022towards}. However, the performance of these methods is still intrinsically limited by the data quality and quantity in both the source and target domains.

Besides model development, text data augmentation can also overcome the sample size limit and work together with other FSL methods in NLP~\cite{wei2019eda, kumar2019closer}. Data augmentation is usually model-agnostic and involves no change to the underlying model architecture, which makes this approach particularly practical and applicable to a wide range of tasks. In NLP, there are several types of data augmentation methods. Traditional text-level data augmentation methods rely on direct operations on the existing sample base. Some frequently used techniques include synonym replacement, random deletion, and random insertion~\cite{feng2021survey}. More recent methods utilize language models to generate reliable samples for more effective data augmentation, including back-translation~\cite{sennrich2015improving} and word vector interpolation in the latent space~\cite{jindal2020augmenting}. However, existing data augmentation methods are limited in the accuracy and diversity of the generated text data, and human annotation is still mandatory in many application scenarios~\cite{feng2021survey, shorten2021text, bayer2022survey}.

The advent of (very) large language models (LLMs) such as the GPT family~\cite{brown2020language, min2021recent} brings new opportunities for generating text samples that resemble human-labeled data~\cite{zhou2023comprehensive}, which significantly alleviates the burden of human annotators~\cite{liu2022survey}. LLMs are trained in self-supervised manners, which scale up with the amount of text corpus available in the open domains. The large parameter space of LLMs also allows them to store a large amount of knowledge, while large-scale pre-training (e.g., the autoregressive objective in training GPTs) enables LLMs to encode rich factual knowledge for language generation even in very specific domains ~\cite{wang2023chatcad}. Furthermore, the training of ChatGPT follows that of Instruct-GPT~\cite{ouyangtraining}, which utilizes reinforcement learning with human feedback (RLHF), thus enabling it to produce more informative and impartial responses to input.

Inspired by the success of language models in text generation, we propose a new data augmentation method named AugGPT, which leverages ChatGPT to generate auxiliary samples for few-shot text classification. We test the performance of AugGPT via experiments on both general domain and medical domain datasets. Performance comparison of the proposed AugGPT approach with existing data augmentation methods shows double-digit improvements in sentence classification accuracy. Further investigation into the faithfulness and compactness of the generated text samples reveals that AugGPT can generate more diversified augmented samples while simultaneously maintaining their accuracy (i.e., semantic similarity to the original labels). We envision that the development of LLMs will lead to human-level annotation performance, thus revolutionizing the field of few-shot learning and other tasks in NLP.


\begin{figure*}[!t]
\centering
\includegraphics[width=6.4in]{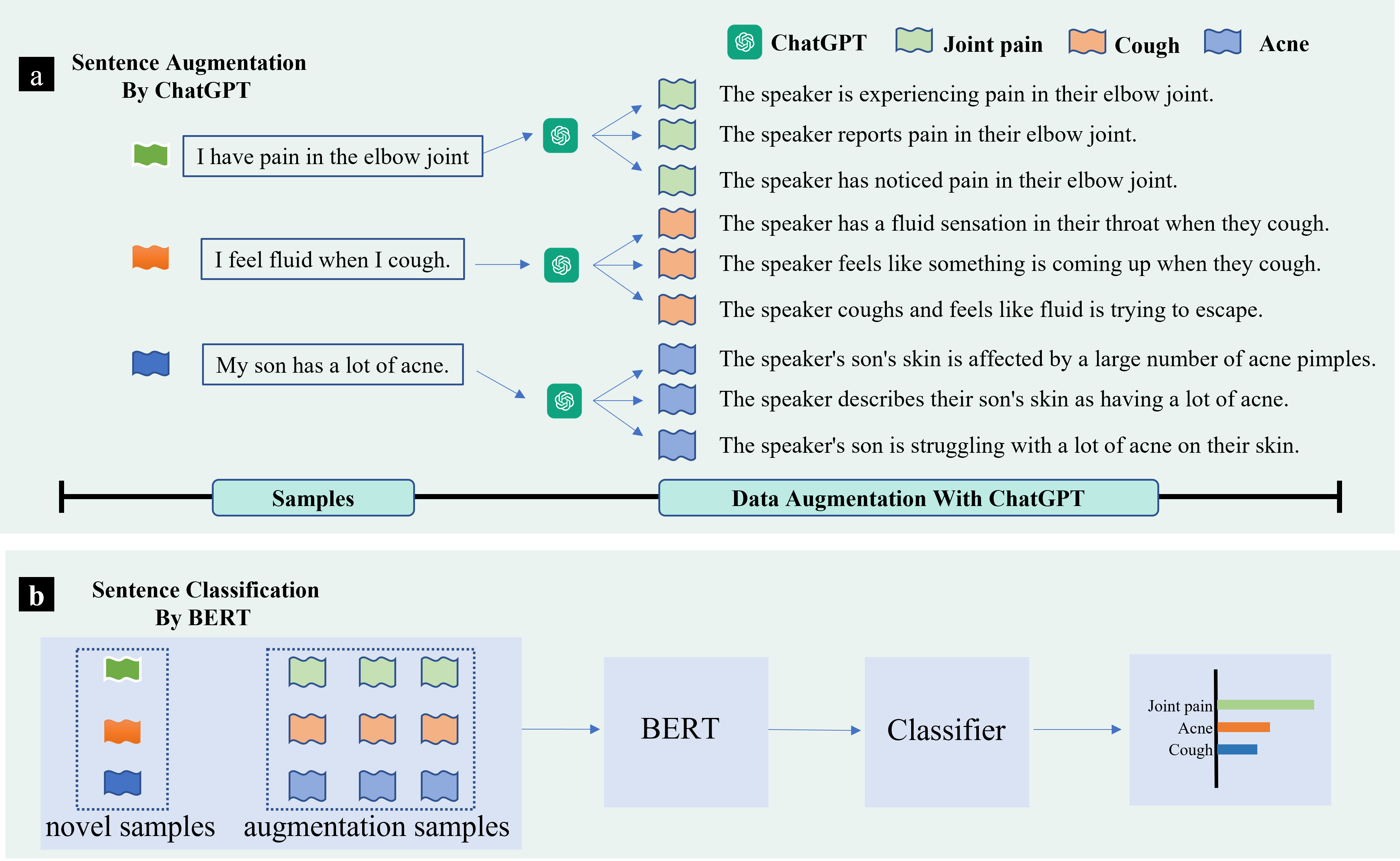}
\caption{The framework of AugGPT. a (top panel): First, we apply ChatGPT for data augmentation. We input samples of all classes into ChatGPT and prompt ChatGPT to generate samples that preserves semantic consistency with existing labelled instance. b (bottom panel): In the next step, we train a BERT-based sentence classifier on the few-shot samples and the generated data samples and evaluate the model's classification performance.}
\label{framework}
\end{figure*}

\begin{table*}[t]
\centering
\caption{ Generated Samples From ChatGPT on PubMed20K}
\label{table1}
{
\begin{tabular}{lll}
\hline
Original Text             & Altered Text                \\ \hline
\multirow{6}{*}{The follow-up rate after 5 years was 85\%.}                                                                & 1. The study showed a follow-up rate of 85\% after 5 years.                 \\
& 2. The 5-year follow-up rate was found to be significantly high at 85\%.    \\
& 3. The participants had an impressive follow-up rate of 85\% after 5 years. \\
& 4. The data indicated a follow-up rate of 85\% 5 years post-study.          \\
& 5. The 5-year follow-up rate was recorded as 85\%, according to the findings. \\
& 6. The results of the study showed a follow-up rate of 85\% after a 5-year period.  \\ \hline
\multirow{5}{*}{\begin{tabular}[c]{@{}l@{}}This means that the maturation of calluses \\ was 27\% faster in the LIPUS group.\end{tabular}}  & 1. The LIPUS group showed a 27\% acceleration in callus maturation.         \\
& \begin{tabular}[c]{@{}l@{}}2. The results revealed that the callus maturation process was 27\% faster in the LIPUS group.\end{tabular}  \\
& \begin{tabular}[c]{@{}l@{}}3. The study found that the callus maturation rate in the LIPUS group was 27\% quicker.\end{tabular}      \\
& \begin{tabular}[c]{@{}l@{}}4. Callus maturation was 27\% more rapid in the LIPUS group compared to others.\end{tabular} \\
& \begin{tabular}[c]{@{}l@{}}5. The callus maturation process in the LIPUS group was 27\% more efficient, according to the \\results.\end{tabular} \\ \hline
\multirow{6}{*}{\begin{tabular}[c]{@{}l@{}}Test Registration: IRCT.ir \\ IRCT2012071010230N1.\end{tabular}}                & \begin{tabular}[c]{@{}l@{}}1. The trial was registered with the IRCT under the identifier IRCT2012071010230N1.\end{tabular}                       \\
& \begin{tabular}[c]{@{}l@{}}2. The study was officially registered with IRCT.ir, with the registration number \\ IRCT2012071010230N1.\end{tabular} \\
& \begin{tabular}[c]{@{}l@{}}3. The trial was documented with IRCT.ir with the identification code IRCT2012071010230N1.\end{tabular}              \\
& \begin{tabular}[c]{@{}l@{}}4. IRCT.ir served as the official registrar for the trial, with the registration number \\IRCT2012071010230N1.\end{tabular}   \\
& \begin{tabular}[c]{@{}l@{}}5. The study was recorded with IRCT under the registration number IRCT2012071010230N1.\end{tabular}  \\
& 6. IRCT.ir recorded the trial with the identifier IRCT2012071010230N1.    \\ \hline
\multirow{6}{*}{\begin{tabular}[c]{@{}l@{}}While behavioral and technological \\ interventions can lead to some modest \\ improvements in glycemic control,  \\ these interventions have not performed \\ much better than conventional prevention \\ in achieving glycemic control.\end{tabular}} & \begin{tabular}[c]{@{}l@{}}1. The study found that although behavioral and technological interventions led to some \\ slight improvements in glycemic control, they were not significantly more effective \\ than typical care.\end{tabular}      \\
& \begin{tabular}[c]{@{}l@{}}2. Despite the modest improvement in glycemic control through behavioral and \\ technological interventions, they did not perform better than the standard care.\end{tabular}                                         \\
& \begin{tabular}[c]{@{}l@{}}3. The results showed that while behavioral and technological interventions resulted \\ in some minimal gains in glycemic control, they did not surpass the usual care in \\ achieving glycemic control.\end{tabular} \\
& \begin{tabular}[c]{@{}l@{}}4. Although behavioral and technological interventions showed some improvement \\ in glycemic control, they were not found to be significantly superior to the usual care.\end{tabular}                               \\
& \begin{tabular}[c]{@{}l@{}}5. The study showed that the usual care was not outperformed by behavioral and \\ technological interventions in terms of achieving glycemic control, despite some \\ small improvements.\end{tabular}                \\

 \hline
\end{tabular}
}
\end{table*}

\section{Related work}
\subsection{Data Augmentation} \label{data_aug_section}
Data augmentation, the artificial generation of new text through transformations, is widely used to improve model training in text classification. In NLP, existing data augmentation methods work at different granularity levels: characters, words, sentences, and documents. 

Data augmentation at the character level refers to the randomly inserting, exchanging, replacing, or deleting of characters in the text ~\cite{belinkov2017synthetic}, which improves the robustness of the NLP model against noises. Another method called optical character recognition (OCR) data augmentation generates new text by simulating the errors that occur when using OCR tools to recognize text from pictures. Spelling augmentation \cite{coulombe_text_2018} deliberately misspells some frequently misspelled words. Keyboard augmentation \cite{belinkov2017synthetic} simulates random typo errors by replacing a selected key with another key close to it on the QWERTY layout keyboard.

Data augmentation also works at the word level. Random swap augmentation randomly exchanges two words in the text, and random deletion augmentation randomly deletes some words~\cite{wei_eda_2019}. Synonym augmentation uses synonym databases such as PPDB~\cite{pavlick2015ppdb} to replace randomly selected words~\cite{niu_adversarial_2018}. WordNet~\cite{miller1995wordnet} is also widely used as a reference for synonym augmentation. These methods maintain semantic consistency and are suitable for text classification tasks. Wang et al.~\cite{wang2015s} proposed a data augmentation method based on word embeddings, replacing words with their top-$n$ similar words to create a new sentence. Different pre-trained word embeddings are considered (e.g., GoogleNews Lexical Embeddings~\cite{mikolov2013distributed}). This method is based on the principle that words close to each other in the embedding space often appear in similar contexts, which might help with maintaining grammatical consistency.

However, a serious limitation of word embedding-based methods is that close words in the embedding space are not necessarily semantically similar, yet semantic changes can affect the classification results. For example, ``hot" and ``cold" usually appear in similar contexts, so their word embeddings are close, but they have exactly opposite semantic meanings. The counter-fitting embedding augmentation ~\cite{mrksic_counter-fitting_2016,alzantot_generating_2018} solves this problem by using a synonym dictionary and an antonym dictionary to adjust the initial word embeddings. Specifically, the distance between embeddings of synonyms will be shortened, and the distance between embeddings of antonyms will become enlarged.

Contextual augmentation ~\cite{kobayashi_contextual_2018,kumar_data_2020} is another word-level data augmentation method, which uses masked language models (MLMs) such as BERT\cite{devlin_bert_2019,sun2020mixup}, DistilBERT\cite{Sanh2019DistilBERTAD} and RoBERTA\cite{liu2019roberta} to generate new text based on the context. Specifically, they insert $<mask>$ tokens in some positions of the text, or replace some words in the text with $<mask>$ tokens, and then let the MLM predict what words should be put in these masked positions. Since MLMs are pre-trained on a large number of texts, contextual augmentation can usually generate meaningful new texts.

Some text data augmentation methods work at the sentence and document level. For example, back translation ~\cite{sennrich_improving_2016} uses translation models for data augmentation. Specifically, the language model first translates the text into another language and then translates it back to the original language. Due to the randomness of the translation process, the augmented text is different from the original text, but semantic consistency is maintained. At the document level, Gangal et al. ~\cite{gangal2022nareor} proposed a method to paraphrase the entire document to preserve document-level consistency. 

In general, regardless of the granularity level or the text generation backbone (i.e., rule-based or language models), the goal of data augmentation is to produce sensible and diverse new samples that maintain semantic consistency. 

\subsection{Few-shot Learning}
Deep learning has achieved remarkable success in various data-intensive applications. However, the performance of deep models could be affected if the dataset size is small in the downstream tasks. Few-shot Learning is a branch of science that focuses on developing solutions to address the challenge of small sample sizes ~\cite{wang2020generalizing,fei2006one}. FSL research aims to leverage prior knowledge to rapidly generalize to new tasks that contain only a few labeled samples. A classic application scenario for few-shot learning is when obtaining supervised examples is difficult or not possible due to privacy, safety, or ethical considerations. The development of few-shot learning enables practitioners to improve the efficiency and accuracy of text classification in various scenarios and deploy practical applications.

Recent advances in few-shot learning have shown promising results in overcoming the challenges of limited training data for text classification. For example, a common approach in NLP is to use a pre-trained language model such as BERT ~\cite{devlin2018bert} as a starting point and then fine-tune it with limited samples. Some of the most recent methodological developments ~\cite{yin2020meta, ge2022few} approaches that have gained traction include prompt-tuning ~\cite{brown2020language, lester2021power, han2022ptr, wang2022towards} and meta-learning ~\cite{yin2020meta, lee2022meta}. In general, existing FSL methods target either architectural design ~\cite{yin2020meta, wang2021transprompt}, data augmentation ~\cite{wei2019eda, kumar2019closer} or the training process~\cite{wei2021few}.

Despite the recent development of prompt-tuning and meta-learning methods, they suffer from some major limitations. For example, prompt engineering is a cumbersome art that requires extensive experience and manual trial-and-errors~\cite{gao2021making}. Meta-learning, on the other hand, suffers from problems such as training instability~\cite{antoniou2018train, finn2017model, yao2021model} and sensitivity to hyper-parameters~\cite{antoniou2018train, finn2017model}. In addition, all these FSL pipelines demand deep machine learning expertise and acquaintance with complex model architectures and training strategies, which are not attainable by common practitioners and general developers. As discussed in section ~\ref{data_aug_section}, data augmentation is an effective solution for FSL and can be combined with other FSL models. Thus, the AugGPT method proposed in this paper, which has demonstrated the capability to generate accurate and comprehensive training samples, can overcome the issues of current FSL methods and potentially change the landscape of few-shot learning in NLP.


\subsection{Very Large Language Models}
Pre-trained language models (PLMs) based on the transformer architecture, such as the BERT ~\cite{devlin2018bert} and GPT~\cite{radford2018improving} model families, have revolutionized natural language processing. Compared to previous methods, they deliver state-of-the-art performance on a wide range of downstream tasks and contribute to the rising popularity and democratization of language models. In general, there are three classes of pre-trained language models: autoregressive language models (e.g., the decoder-based GPT), masked language models (e.g., the encoder-based BERT), and encoder-decoder models(e.g., BART~\cite{lewis2019bart} and T5 ~\cite{raffel2020exploring}). These models typically contain between 100M and 1B parameters ~\cite{min2021recent}.

In recent years, NLP communities have witnessed the rise of very large language models such as GPT-3 (175B parameters)~\cite{brown2020language}, PaLM (540B parameters)~\cite{chowdhery2022palm}, Bloom (176B parameters)~\cite{scao2022bloom}, OPT (up to 175B parameters)~\cite{zhang2022opt}, and the FLAN series (FLAN has 137B parameters)~\cite{longpre2023flan}. At their core, these large language models are transformer models inspired by BERT and GPT, albeit at a much larger scale. 

Large language models aim to learn accurate latent feature representations of input text. These representations are often context-dependent and domain-dependent. For example, the vector representation of the word "treat" might be vastly different between medical domains and the general domain. For smaller pre-trained language models, it is often necessary to continuously pre-train and fine-tune such models to attain acceptable performance~\cite{gu2021domain}. However, very large language models can potentially eliminate the need for fine-tuning while maintaining competitive performance~\cite{brown2020language,zhou2023comprehensive,rezayi2022clinicalradiobert}. 

Existing studies indicate that pre-trained language models can help augment a dataset with new samples with similar semantic meaning~\cite{feng2021survey,bayer2022survey}, which is of significant practical value to real-world applications. In this study, we aim to use ChatGPT, a popular LLM to conduct data augmentation. ChatGPT is based on GPT-3 ~\cite{brown2020language}, which was trained on massive web data with diverse and rich information. Furthermore, ChatGPT was trained through Reinforcement learning from Human Feedback (RLHF). During RLHF, human feedback is incorporated into the process of generating and selecting the best results. More specifically, a reward model is trained based on human annotators' ranking or generated results. In turn, this reward model rewards model outputs that are most aligned with human preference and human values. We believe these innovations make ChatGPT the best candidate for generating human-level quality data samples. 

\subsection{ChatGPT: Present and Future}
ChatGPT is a game changer in natural language processing. For the first time in human history, the power of large language models is accessible to the general public through a user-friendly chatbot interface. In turn, this common accessibility contributes to ChatGPT's unprecedented popularity. ChatGPT has emerged as a general-purpose problem solver for many NLP applications~\cite{qin2023chatgpt}. Qin et al.~\cite{qin2023chatgpt} evaluated ChatGPT on a comprehensive set of NLP tasks, including common benchmarks in natural language inference, arithmetic reasoning, named entity recognition, sentiment analysis, question answering, dialogue and summarization. They conclude that ChatGPT excels in most tasks, except for tasks that focus on specific details (e.g., sequence tagging). 

ChatGPT is also a valuable solution for multilingual tasks. A recent empirical study~\cite{jiao2023chatgpt} reports that ChatGPT excels at tasks involving high-resource languages (various European languages and Chinese) and is comparable with Google Translate, DeepL Translate and Tencent TranSmart. Nonetheless, ChatGPT performs poorly on low-resource languages and faces extra challenges handling distant language translation (i.e., English-German translation is considered to be less "distant", compared to English-Hindi translation). A later study~\cite{bang2023multitask} confirms that ChatGPT struggles with low-resource languages, although the authors observe that ChatGPT does better in understanding non-Latin scripts than generating them. 

In addition, it is also possible to use the purely text-based ChatGPT to interact with multi-modal data. A group of researchers ~\cite{bang2023multitask} use HTML Canvas and Python Turtle graphics as media for text-to-image generation. ChatGPT can faithfully generate HTML and Python code, which can be then used to generate desired images. The authors designed a flag drawing task that required ChatGPT to generate code that can generate country flags. It was found that ChatGPT could generate better flags when the prompt for code was preceded by a prompt that queries ChatGPT for the flag's description. In other words, descriptive text prompts could improve multimodal task performance. 

Beyond computer science, ChatGPT can be readily applied to medical report generation and comprehension~\cite{shen2023chatgpt, antaki2023evaluating}, education ~\cite{kung2023performance,pavlik2023collaborating,baidoo2023education}, rigorous math research~\cite{frieder2023mathematical} and finance ~\cite{dowling2023chatgpt}. Overall, ChatGPT is a versatile tool that promotes general AI usage. 

However, researchers are also cautious about the possible negative impact of ChatGPT. Some of the more prominent concerns are related to bias~\cite{van2023chatgpt,mcgee2023chat}, ethics~\cite{blum2022breaking,jabotinsky2022co}, plagiarism~\cite{susnjak2022chatgpt, khalil2023will} and job replacement \textit{en masse}~\cite{castelvecchi2022chatgpt,zarifhonarvar2023economics}. In response, a commentary published in Nature advocates for urgent attention to accountability, open-source large language models and societal embrace of AI~\cite{van2023chatgpt}. 

\section{Dataset}
We first use an open domain dataset Amazon to verify the effectiveness of our method. Then, we use clinical natural language processing (clinical NLP) as the task and carry out our experiments on two popular public benchmarks. Data augmentation is particularly in demand in clinical NLP, because the significant burden of expert annotation and stringent privacy regulations make large-scale data labeling infeasible. We will describe these datasets in detail in the following sections. 

\subsection{Amazon dataset}
Amazon\cite{he2016ups,bao2019few,wang2022sentence} contains customer reviews from 24 product categories.  The task is to classify reviews into their respective product categories. Since the original Amazon product dataset is proverbially large, we sample a subset of 300 samples from each category.

\subsection{Symptoms Dataset}
This dataset is published on Kaggle\footnote{https://www.kaggle.com/datasets/paultimothymooney/medical-speech-transcription-and-intent}. It contains the audio data of common medical symptom descriptions over 8 hours. We use the text transcripts corresponding to the audio data and perform sample de-duplication, and use them as model input. The dataset after preprocessing includes 231 samples of 7 symptom categories. Every example represents a sentence describing the provided symptoms, and the task is to classify the sentence into the corresponding symptoms.

\subsection{PubMed20k Dataset}
The PubMed20K dataset is an extensively utilized resource in NLP and text mining research, comprising around 20,000 annotated scientific abstracts from the biomedical field. These annotations encompass named entities, relationships between entities, and various semantic roles, making the dataset valuable for diverse NLP tasks such as named entity recognition, relation extraction, and text classification. The dataset originates from the PubMed database, which spans a wide array of biomedical subjects. Owing to its substantial size, variety, and high-quality annotations, PubMed20K has emerged as a popular benchmark dataset for assessing the performance of machine learning models in the realm of biomedical NLP. The abstracts in the PubMed 20K dataset undergo preprocessing and segmentation into individual sentences. Each sentence is labeled with one of the following five categories: background, objective, method, result, or conclusion. The task is to map the input sentences to their corresponding categories.

\section{Method}


\subsection{Overall Framework}

\begin{algorithm}[tb]
\caption{The framework of AugGPT for few-shot text classification.}
\label{algorithm1}
\textbf{Input}: base dataset ${D}_b$ and novel dataset ${D}_n$  \\
\textbf{Initialize}: Initialized pre-trained BERT $model$ \\
\textbf{Definition}: $D^{'}$ is the dataset with the base dataset ${D}_b$ and augmented dataset $D_n^{aug}$, and \textit{chatGPT\_aug} is the data augmentation method based on ChatGPT \\
    \textbf{Parameters}: Fine-tuning epochs of base dataset $epoch_b$, fine-tuning epochs of  FSL $epoch_f$ 
\begin{algorithmic}
    \FOR{epoch {\bfseries in} $epoch_b$}
        \STATE train($model$, $D_b$)
    \ENDFOR
    \STATE $D_n^{aug}$ = \textit{chatGPT\_aug}(${D}_n$)
    \FOR{epoch {\bfseries in} $epoch_f$}
        \STATE train($model$,$D_n^{aug}$)
    \ENDFOR
\end{algorithmic}
\end{algorithm}

Given a base dataset $D_b = \{(x_i,y_i)\}_{i=1}^{N_{b}}$ with a label space $y_i \in Y_b$, a novel dataset $D_n = \{(x_j,y_j)\}_{j=1}^{N_{n}}$ with a label space $y_j \in Y_n$, and $Y_b \cap Y_n = \emptyset$. In the few-shot classification scenario, the base dataset $D_b$ has a relatively larger set of labeled samples, while the novel dataset $D_n$ has only a few labeled samples. The performance of few-shot learning is evaluated on the novel dataset. Our goal is to train a model with both base and limited novel datasets, while achieving satisfying generalizability on the novel dataset. 

The overall framework of AugGPT is shown in Fig~\ref{framework}, and the training steps are shown in Algorithm~\ref{algorithm1}. First of all, we fine-tune BERT on $D_b$. Then, the $D_n^{aug}$ is generated by data augmentation with ChatGPT. Finally, we fine-tune BERT with $D_n^{aug}$.

\subsection{Data Augmentation with ChatGPT}

Similar to GPT \cite{radford2018improving}, GPT-2 \cite{radford2019language}, and GPT-3 \cite{brown2020language}, ChatGPT belongs to the family of autoregressive language models and uses transformer decoder blocks \cite{vaswani2017attention} as the model backbone.

During pre-training, ChatGPT is regarded as an unsupervised distribution estimation from a set of samples $X = \{x_1,x_2,...,x_n\}$, and sample $x_i$ composed of $m$ tokens is defined as $x_i = (s_1,s_2,...,s_m)$. The objective of pre-training is to maximize the following likelihood:
\begin{equation}\label{eq1}
L(x_i) = \sum_{i=1}^{m} \log{P(s_i|s_{1},...,s_{i-1};\theta)}
\end{equation}
where $\theta$ represents the trainable parameters of ChatGPT.
The tokens are represented by token embedding and position embedding:
\begin{equation}\label{eq2}
h_0 = x_iW_e + W_p
\end{equation}
where $W_e$ is the token embedding matrix and $W_p$ is the position embedding matrix.
Then $N$ transformer blocks are used to extract the features of the sample:
\begin{equation}\label{eq3}
h_n = transformer\_blocks(h_{n-1})
\end{equation}
where $n \in [1,N]$.

Finally, the target token is predicted:
\begin{equation}\label{eq4}
s_i = softmax(h_N W_e^T)
\end{equation}
where $h_N$ is the output of top transformer blocks.

After pre-training, the developers of ChatGPT apply Reinforcement Learning from Human Feedback (RLHF)\cite{ouyangtraining} to fine-tune the pre-trained language model. The RLHF aligns language models with user intent on a wide range of tasks by fine-tuning them according to human feedback. The RLHF of ChatGPT contains three steps:

\textbf{Supervised Fine-tuning (SFT)}: Unlike GPT, GPT-2, and GPT-3, ChatGPT uses labeled data for further training. The AI trainers play as users and AI assistants to build the answers based on prompts. The answers with prompts are used as supervised data for further training of the pre-trained model. After further pre-training, a SFT model can be obtained.

\textbf{Reward Modeling (RM)}: Based on the SFT method, a reward model is trained to take in a pair of prompt and response, and output a scalar reward. Human labelers rank the outputs from best to worst to build a ranking dataset. The loss function between two outputs is defined as follows:
\begin{equation}\label{eq5}
\operatorname{loss}(\theta_{r})= E_{\left(x, y_w, y_l\right) \sim D_c}\left[\log \left(\sigma\left(r_{\theta_{r}}\left(x, y_w\right)-r_{\theta_{r}}\left(x, y_l\right)\right)\right)\right]
\end{equation}
where $\theta_{r}$ is the parameters of reward model; $x$ is the prompt, $y_w$ is the preferred completion out of the pair of $y_w$ and $y_l$; $D_c$ is the dataset of human
comparisons.

\textbf{Reinforcement Learning (RL)}: By using reward models, ChatGPT can be fine-tuned using Proximal Policy Optimization (PPO) \cite{schulman2017proximal}. In order to fix the performance degradation on public NLP datasets, the RLHF mixes the pretraining gradients into the PPO gradients, which is also known as PPO-ptx:
\begin{equation}
\label{eq6}
\begin{aligned}
& \operatorname{objective}(\phi)= \gamma E_{x \sim D_{\text {pretrain}}}\left[\log \left(\pi_\phi^{\mathrm{RL}}(x)\right)\right] + \\
& E_{(x, y) \sim D_{\pi_\phi^{\mathrm{RL}}}}\left[ r_{\theta_{r}}(x, y)- \beta \log \left(\pi_\phi^{\mathrm{RL}}(y \mid x) / \theta_{\mathrm{SFT}}(y \mid x)\right) \right] 
\end{aligned}
\end{equation}
where $\pi_\phi^{\mathrm{RL}}$ is the learned RL policy, $\theta_{\mathrm{SFT}}$ is the supervised trained model, and $D_{\text {pretrain}}$ is the pretraining distribution. The $\gamma$ is the pre-training loss coefficient that controls the strength of pre-training gradients, and the $\beta$ is the KL (Kullback-Leibler) reward coefficient that controls the strength of the KL penalty.

Compared to previous data augmentation methods, ChatGPT is more suitable for data augmentation for the following reasons:
\begin{itemize}
    \item ChatGPT is pre-trained on large-scale corpora, so it has a broader semantic expression space, and is helpful to enhance the diversity of data augmentation.
    \item Since the fine-tuning stage of ChatGPT introduces a large number of manual annotation samples, the language generated by ChatGPT is more in line with human expression habits.
    \item Through reinforcement learning, ChatGPT can compare the advantages and disadvantages of different expressions and ensure that the generated data are of high quality.
\end{itemize}

Under the BERT framework, we introduce ChatGPT as the data augmentation tool for few-shot text classification. Specifically, ChatGPT is applied to rephrase each input sentence into six additional sentences, thereby augmenting the few-shot samples.

\subsection{Few-shot Text Classification}
We apply BERT~\cite{devlin2019bert} to train a few-shot text classification model. The output features $h$ of the top layer of BERT can be written as:
\begin{equation}
\label{eq7}
z = [z_{c},z_1,z_2,...,z_n],
\end{equation}
where $z_{c}$ is the representation of the class-specific token CLS. For text classification, $z_{c}$ is usually fed into a task-specific classifier header for final prediction. However, in the FSL scenario, it is difficult to achieve satisfactory performance through BERT fine-tuning because the small scale of few-shot samples will easily lead to over-fitting and lack of generalization ability.

To effectively address the challenge of few-shot text classification, many approaches have been proposed. Generally, there are four categories of methods for few-shot text classification based on large language models: meta-learning, prompt-tuning, model design, and data augmentation. meta-learning refers to the process of \textit{learning to learn} with tasks that update meta-parameters ~\cite{yin2020meta, lee2022meta}. Prompt-based methods guide large language models to predict correct results by designing templates ~\cite{brown2020language, lester2021power, han2022ptr, wang2022towards}. Model design methods guide the model to learn from few-shot samples by changing the structure of the model \cite{mask_bert2023liao}. Data augmentation uses similar characters \cite{belinkov2017synthetic}, similar word semantics \cite{ mrksic_counter-fitting_2016,alzantot_generating_2018}, or knowledge base \cite{rezayiagribert, rezayi2022clinicalradiobert} to expand samples. Our method directly data augmentation through the language capabilities of large language models, which is a simple and efficient data augmentation method.

\textbf{Objective Function}: Our objective function of few-shot learning consists of two parts: cross entropy and contrastive learning loss. We feed $z_{c}$ into a fully connected layer, the classifier for the final prediction:
\begin{equation}\label{eq8}
\hat{y} = W_c^T z_{c} + b_c,
\end{equation}
where $W_c$ and $b_c$ are trainable parameters, and take cross-entropy as one of the objective functions:
\begin{equation}\label{eq9}
L_{CE} = - \sum_{d\in D' } \sum_{c=1}^C y_{dc} \ln \hat{y}_{dc},
\end{equation}
where $C$ is the output dimension, which is equal to the union of label spaces of the base dataset and novel dataset, and $y_d$ is the ground truth.

Then, to make full use of the prior knowledge in the base dataset to guide the learning of the novel dataset, we introduce the contrastive loss function to make the sample representation of the same category more compact and the sample representation of different categories more separate. The contrastive loss between pairs of samples in the same batch is defined as follows:
\begin{align}\label{eq10}
L_{CL} =  - \log \frac{{\sum {e^{cos({v_i},{v_{i'}})}} }}{{\sum {e^{cos({v_i},{v_{i'}})}}  + \sum {e^{cos({v_i},v_j)}} }},
\end{align}
where $v_i$ and $v_i^{'}$ are the $z_{c}$ of samples that belong to the same category; $v_i$ and $v_j$ are the $z_{c}$ of samples belong to different categories; $cos(\cdot;\cdot)$ is the cosine similarity.

In the BERT fine-tuning stage on the base dataset, we only use cross entropy as the objective function. In the few-shot learning stage, we combine cross entropy and contrastive learning loss as the objective function:
\begin{align}\label{eq11}
L =  L_{CE} + \lambda L_{CL} .
\end{align}

\begin{table*}[t]
\centering
\caption{ Data Augmentation and Ablation Study. The  BERT + C indicates BERT with contrastive loss. }
\label{table2}
{
\begin{tabular}{llclclc}
\hline
\multirow{2}{*}{Data Augmentation}           & \multicolumn{2}{c}{Amazon}      & \multicolumn{2}{c}{Symptoms}    & \multicolumn{2}{c}{PubMed20K}   \\ \cline{2-7} 
                                             & BERT           & BERT + C       & BERT           & BERT + C       & BERT           & BERT + C       \\ \hline
Raw                                          & 0.734          & 0.745          & 0.636          & 0.606          & 0.792          & 0.798          \\
BackTranslationAug                           & 0.757          & 0.748          & 0.778          & 0.747          & 0.812          & 0.83           \\
ContextualWordAugUsingBert(Insert)           & 0.761          & 0.750          & 0.697          & 0.677          & 0.802          & 0.811          \\
ContextualWordAugUsingBert(Substitute)       & 0.770          & 0.757          & 0.626          & 0.667          & 0.815          & 0.830           \\
ContextualWordAugUsingDistilBERT(Insert)     & 0.759          & 0.762          & 0.707          & 0.747          & 0.796          & 0.796          \\
ContextualWordAugUsingDistilBERT(Substitute) & 0.787          & 0.766          & 0.667          & 0.646          & 0.797          & 0.800            \\
ContextualWordAugUsingRoBERTA(Insert)        & 0.775          & 0.768          & 0.758          & 0.707          & 0.815          & 0.814          \\
ContextualWordAugUsingRoBERTA(Substitute)    & 0.745          & 0.730          & 0.727          & 0.667          & 0.782          & 0.782          \\
CounterFittedEmbeddingAug                    & 0.754          & 0.741          & 0.667          & 0.626          & 0.805          & 0.805          \\
InsertCharAugmentation                       & 0.771          & 0.775          & 0.404          & 0.475          & 0.826          & 0.831          \\
InsertWordByGoogleNewsEmbeddings             & \textbf{0.816} & 0.794          & 0.636          & 0.677          & 0.786          & 0.784          \\
KeyboardAugmentation                         & 0.764          & 0.766          & 0.545          & 0.505          & 0.809          & 0.815          \\
OCRAugmentation                              & 0.775          & 0.782          & 0.768          & 0.778          & 0.789          & 0.789          \\
PPDBSynonymAug                               & 0.691          & 0.690          & 0.697          & 0.758          & 0.795          & 0.829          \\
SpellingAugmentation                         & 0.727          & 0.736          & 0.697          & 0.707          & 0.808          & 0.811          \\
SubstituteCharAugmentation                   & 0.762          & 0.768          & 0.535          & 0.586          & 0.816          & 0.821          \\
SubstituteWordByGoogleNewsEmbeddings         & 0.729          & 0.741          & 0.727          & 0.727          & 0.807          & 0.822          \\
SwapCharAugmentation                         & 0.762          & 0.766          & 0.475          & 0.485          & 0.797          & 0.801          \\
SwapWordAug                                  & 0.771          & 0.766          & 0.687          & 0.727          & 0.798          & 0.794          \\
WordNetSynonymAug                            & 0.805          & 0.798          & 0.616          & 0.758          & 0.761          & 0.757          \\
\hline
ChatGPT (2-shot)                             &\multicolumn{2}{c}{0.753}          &\multicolumn{2}{c}{0.980}             &\multicolumn{2}{c}{0.748} \\
\hline
AugGPT                                      & \textbf{0.816} & \textbf{0.826} & \textbf{0.889} & \textbf{0.899} & \textbf{0.835} & \textbf{0.835} \\ \hline
\end{tabular}
}
\end{table*}




\subsection{Baseline Methods}

In the experiment section, we compare our method with other popular data augmentation methods. For these methods, we use the implementation in open-source libraries including, nlpaug~\cite{ma2019nlpaug} and textattack~\cite{morris2020textattack}. 

\begin{itemize}
    \item \textbf{InsertCharAugmentation}. This method inserts random characters at random locations in text, which improves the generalization ability of the model by injecting noise into the data.
    \item \textbf{SubstituteCharAugmentation}. This method randomly replaces selected characters with other ones.
    \item \textbf{SwapCharAugmentation}~\cite{belinkov2017synthetic}. This method randomly exchanges two characters.
    \item \textbf{DeleteCharAugmentation}. This method randomly deletes characters.
    \item \textbf{OCRAugmentation}. OCRAugmentation simulates possible errors during OCR recognition. For example, OCR tool may wrongly identify ``0" as ``o", and wrongly identify ``I" as ``l".
    \item \textbf{SpellingAugmentation}~\cite{coulombe_text_2018}. It creates new text by deliberately misspelling some words. The method uses a list of English words that are most likely to be misspelled provided by Oxford Dictionary, for example, misspelling ``because" as ``becouse".
    \item \textbf{KeyboardAugmentation}~\cite{belinkov2017synthetic}. It simulates typo error by replacing randomly selected characters with the adjacent characters in the QWERTY layout keyboard. For example, replacing `g' with `r', `t', `y', `f', `h', `v', `b' or `n'.
    \item \textbf{SwapWordAug}~\cite{wei_eda_2019}. It randomly exchanges words in text. This method is a submethod of Easy Data Augmentation (EDA) proposed by Wei et al.
    \item \textbf{DeleteWordAug}. DeleteWordAug randomly deletes words in the text, which is also a submethod of EDA.
    \item \textbf{PPDBSynonymAug}~\cite{niu_adversarial_2018}. It replaces words with their synonym in PPDB thesaurus. Synonym replacement can ensure semantic consistency and is suitable for classification tasks. 
    \item \textbf{WordNetSynonymAug}. It replaces words with their synonym in WordNet thesaurus.
    \item \textbf{SubstituteWordByGoogleNewsEmbeddings}~\cite{wang2015s}. It replaces words with their top-$n$ similar words in the embedding space. The word embeddings used are pre-trained with GoogleNews corpus.
    \item \textbf{InsertWordByGoogleNewsEmbeddings}~\cite{ma2019nlpaug}. It randomly selects word from vocabulary of GoogleNews corpus and inserts it the random position of the text.
    \item \textbf{CounterFittedEmbeddingAug}~\cite{mrksic_counter-fitting_2016,alzantot_generating_2018}. It replaces words with their neighbors in counter-fitting embedding space. Compared with GoogleNews word vectors used by SubstituteWordByGoogleNewsEmbeddings, counter-fitting embedding introduces the constraint of synonyms and antonyms, that is, the embedding between synonyms will be pulled closer, and vice versa.
    \item \textbf{ContextualWordAugUsingBert(Insert)}~\cite{kobayashi_contextual_2018,kumar_data_2020}. This method uses BERT to insert words based on context, that is, add $<mask>$ token at random position of the input text, and then let BERT predict the token at that position.
    \item \textbf{ContextualWordAugUsingDistilBERT(Insert)}. This method uses DistilBERT to replace BERT for prediction, and the rest is the same as ContextualWordAugUsingBert(Insert).
    \item \textbf{ContextualWordAugUsingRoBERTA(Insert)}. This method uses RoBERTA to replace BERT for prediction, and the rest is the same as ContextualWordAugUsingBert(Insert).
    \item \textbf{ContextualWordAugUsingBert(Substitute)}. This method \cite{kobayashi_contextual_2018,kumar_data_2020} uses BERT to replace words based on context, that is, replace randomly selected words in text with $<mask>$ token, and then let BERT predict the token at that position.
    \item \textbf{ContextualWordAugUsingDistilBERT(Substitute)}. This method uses DistilBERT to replace BERT for prediction, and the rest is the same as ContextualWordAugUsingBert(Substitute).
    \item \textbf{ContextualWordAugUsingRoBERTA(Substitute)}. This method uses RoBERTA to replace BERT for prediction, and the rest is the same as ContextualWordAugUsingBert(Substitute).
    \item \textbf{BackTranslationAug}. The method \cite{sennrich_improving_2016} translates the text into German and then into English, resulting in a new text that is different from the original but has the same semantics. We use wmt19-en-de and facebook/wmt19-de-en language translation models \cite{WMT19_facebook} developed by Facebook for translation.
\end{itemize}

\subsection{Prompt Design}
\begin{figure*}[!t]
\centering
\includegraphics[width=7in]{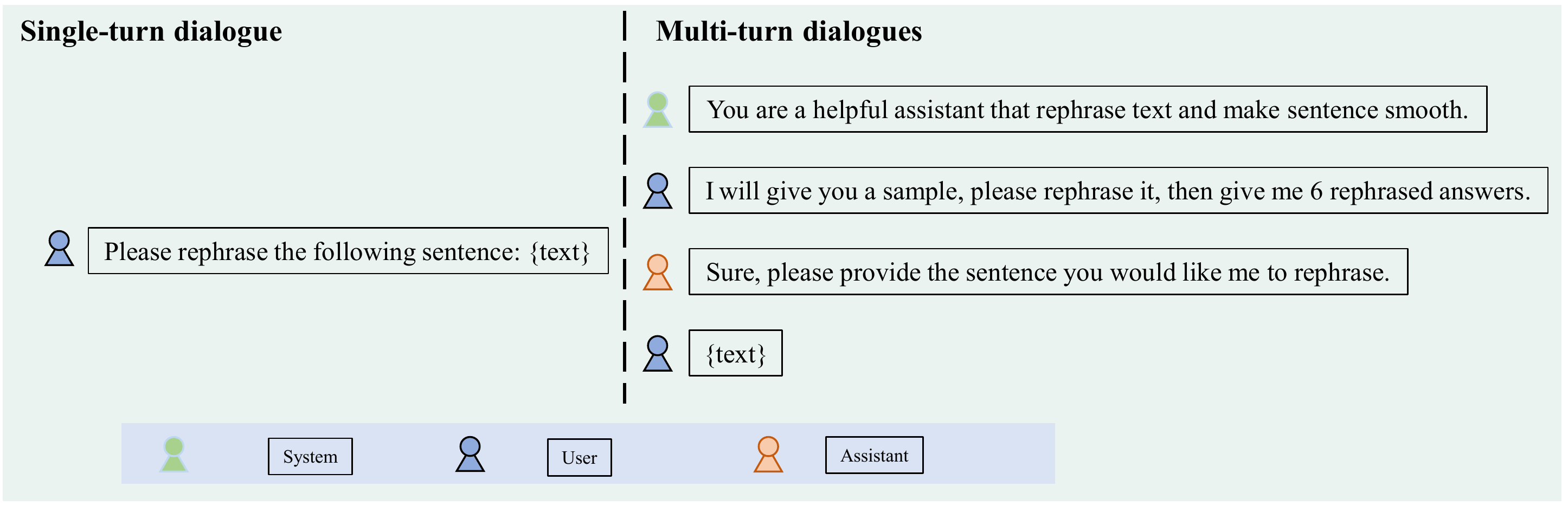}
\caption{ Single-turn dialogue and multi-turn dialogues prompt}
\label{prompt_fig}
\end{figure*}

We have designed prompts for single-turn dialogue and multi-turn dialogues. The prompts are shown in Fig \ref{prompt_fig}. The Amazon dataset use the multi-turn dialogues prompt for data augmentation. The  Symptoms and PubMed20K use the single-turn dialogue prompt for data augmentation. 

\begin{figure*}[!t]
\centering
\includegraphics[width=7in]{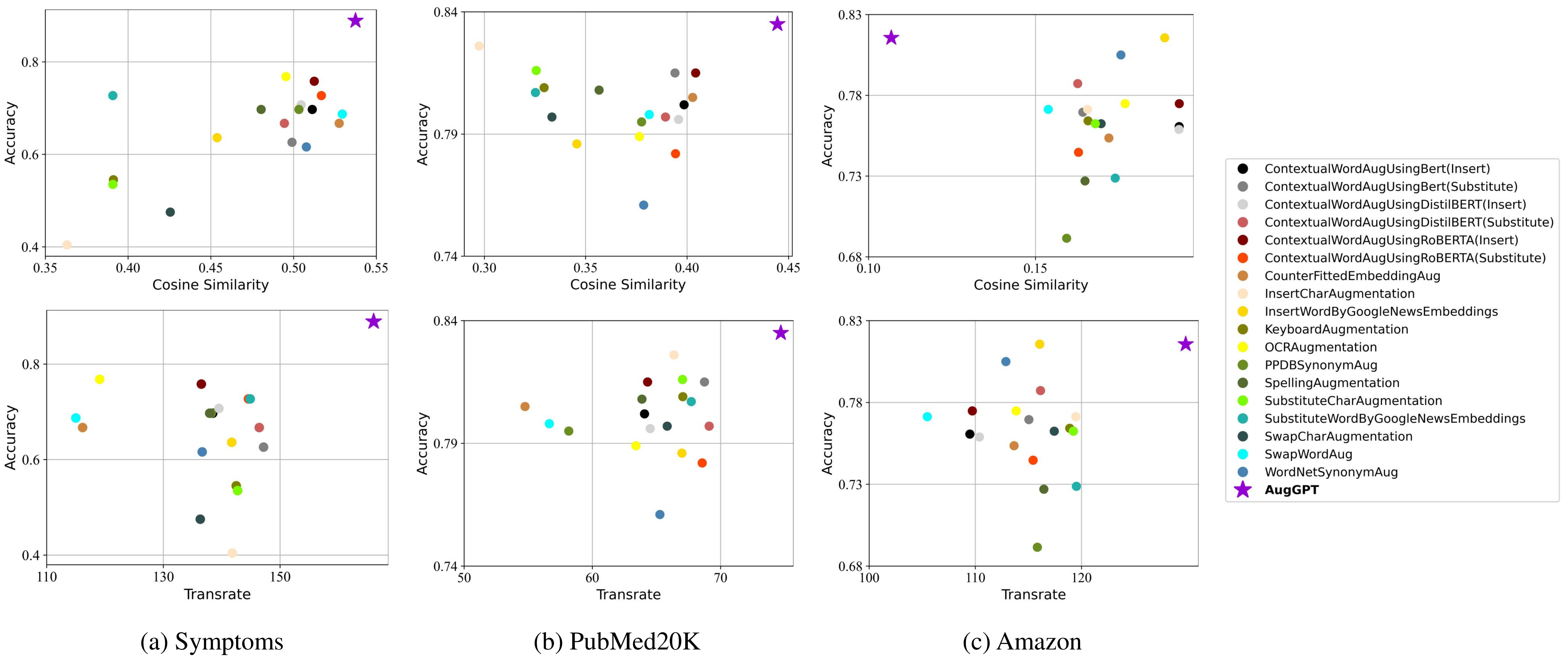}
\caption{We employed two evaluation metrics to assess the faithfulness and compactness of our newly augmented data. The top left plot displays the cosine similarity metric and final accuracy of all data augmentation methods on the Symptoms dataset, and the bottom left plot shows the TransRate metric and final accuracy of all data augmentation methods on the Symptoms dataset. In the middle and bottom panels, we plotted the cosine similarity and TransRate values of all data augmentation methods on the Amazon and PubMed20K datasets, respectively. On the right side of the picture, we listed all the augmented methods with different colors and shapes. }
\label{evaluation_fig}
\end{figure*}

\subsection{Evaluation Metrics} \label{eval_metrics_section}
We employed cosine similarity and TransRate\cite{huang2022frustratingly} as metrics to assess the faithfulness (i.e., whether the generated data samples are close to the original samples) and compactness (i.e., whether samples of each class are compact enough for good discrimination) of the augmented data.

\subsubsection{Embedding Similarity}
To evaluate the semantic similarity between the samples generated by data augmentation methods and actual samples, we adopt embedding similarity between the generated samples and the actual samples of the test dataset. Some of the most common similarity metrics include Euclidean distance, cosine similarity and dot product similarity. In this study, we select cosine similarity to capture the distance relationship in the latent space. The cosine similarity measures the cosine value of the angle between two vectors. This value increases when two vectors are more similar, and is bounded by a range between 0 and 1. Since the pre-trained language models without fine-tunning poorly to capture semantic meaning, we fine-tunning the pre-trained BERT on base dataset by BERT-flow \cite{li2020sentence} method, and  finally apply the fine-tunned BERT to get smaple embedding. The cosine similarity metric is commonly used in NLP ~\cite{wang2020measurement} and we follow this convention. 

\begin{align}\label{eq12}
\cos (\theta) = \dfrac {A \cdot B} {\left\| A\right\| _{2}\left\| B\right\| _{2}},
\end{align}
where A and B denote the two embedding vectors in comparison, respectively. 

\subsubsection{TransRate}
TransRate is a metric that quantifies transferability based on the mutual information between the features extracted by a pre-trained model and their labels, with a single pass through the target data. The metric achieves a minimum value when the data covariance matrices of all classes are identical, making it impossible to distinguish between the data from different classes and preventing any classifier from achieving better than random guessing. Thus, a higher TransRate could indicate better learnability of the data. More specifically, knowledge transfer from a source task $T_s$ to a target task $T_t$ is measured as shown below:
\begin{align}\label{eq13}
TrR_{T_s \to T_t}(g) =  H(Z) - H(Z|Y),
\end{align}
where Y represents the labels of augmented examples, and $Z$ denotes the latency embedding features extracted by the pre-trained feature extractor $g$. $TrR$ means the TransRate value. $H(\cdot)$ denotes the Shannon entropy\cite{cover1999elements}.


\subsection{Direct Classification Performance by ChatGPT}
An interesting and important question about the utilization of ChatGPT for text data augmentation would be how ChatGPT will perform when directly applied to FSL downstream tasks. Thus, we developed tailored prompts for ChatGPT to perform the classification tasks with integrated the API for prompting. For the Symptoms dataset, we employed the following prompt instruction: "Given a person's health description or symptom, predict the corresponding illness from the following categories: {CLASSES}." Additionally, we used "Description: {DESCRIPTION}. Typically, this symptom corresponds to {CLASS}" as the prompt for each example in the dataset. In this way, We can include few-shot examples (in this work, we used two) to facilitate the model's adaptation to downstream tasks. We used similarly-designed prompt instructions for the other two tasks and the corresponding example prompt to implement the few-shot in-context learning by ChatGPT.

\section{Experiment Results}
In our experiments, we use BERT as the base model. Firstly, we train our model on the base dataset to produce the pre-trained model. Then we fine-tune the model with the combination of few-shot samples and the augmented samples generated from various data augmentation methods. Specifically, in all three FSL tasks, we perform 2-shot learning, i.e., there would be two real samples used for each class in the target domain. Afterward, We use those samples to fine-tune the pre-trained models. To evaluate the effectiveness of different data augmentation methods, we apply two different settings. The first one is the vanilla BERT model. In the second setting, we add a contrastive loss to the training objective function. In our experiments on the Symptoms dataset, we use a batch size of 8 for 150 epochs, set the maximum sequence length to 25, $\lambda$ as 1, and use a learning rate of 4e-5. In our experiments on the PubMed20K dataset, we adopt the same training configuration, with the maximum sequence length set to 40. For all three tasks, we will generate six augmented samples per class. Examples of the augmented samples generated by AugGPT and other selected baseline methods can be found in the appendix.

\subsection{Classification Performance Comparison}
Table \ref{table2} shows the accuracy of different data augmentation methods. As shown in Table \ref{table2}, AugGPT achieves the highest accuracy for Amazon,  Symptoms and PubMed20K datasets. For the Amazon dataset, AugGPT and InsertWordByGoogleNewsEmbeddings achieve the best performance for BERT, and AugGPT achieve the best performance for BERT with contrastive loss. In the PubMed20K dataset, AugGPT achieves 83.5\% accuracy for both BERT and BERT with contrastive loss, whereas without data augmentation, the accuracy values are only 79.2\% and 79.8\%, respectively. For the Symptoms dataset, the accuracy for BERT downstream augmentation is only 63.6\%, and 60.6\% with contrastive loss. However, our AugGPT approach significantly improves the accuracy to 88.9\% and 89.9\%, respectively. These results suggest that data augmentation using ChatGPT is more effective in enhancing the performance of machine learning models in various applications.

\subsection{Evaluation of Augmented Datasets}
In addition to the classification accuracy, we evaluate the augmented data in the latent space and visualize the results in Fig \ref{evaluation_fig}. Latent embeddings are evaluated using cosine similarity and the TransRate metric (see section \ref{eval_metrics_section} for more details). The horizontal axis represents the cosine similarity values and Transrate values, and the vertical axis describes the classification accuracy. Since embedded similarity measures the similarity between the generated data and the test dataset, high similarity means that the generated data are close to real input data and with higher faithfulness and compactness. Higher TransRate indicates better learnability of the data. Therefore, a higher TransRate score indicates that the augmented data are of higher quality. The most ideal candidate method should be positioned at the top-right of the visualization. As shown in  Fig \ref{evaluation_fig}, AugGPT produces high-quality samples in terms of both faithfulness and compactness on the Symptoms dataset and the PubMed20K dataset. On the open-domain Amazon dataset, AugGPT also produces high-quality samples with a higher TransRate.

\subsection{Performance Comparison with ChatGPT}
Furthermore, we used ChatGPT to directly perform the downstream text data classification tasks under
a 5-shot learning scheme. We used in-house designed instructions with few-shot in-context examples to prompt ChatGPT as described in 4.7. The performance of ChatGPT for the downstream tasks is listed in Table 2. The result reveals that state-of-the-art large language models such as ChatGPT tend to perform better on relatively easier tasks, for example, identifying symptoms according to a one-sentence description. However, when it comes to complicated tasks such like PubMed, model fine-tuning is still needed and could achieve better performance compared to few-shot prompts.

\section{Conclusion and Discussion}
In this paper, we proposed a novel data augmentation approach for few-shot classification. Unlike other methods, our model expands the limited data at the semantic level to enhance data consistency and robustness, which results in a better performance than most of the current text data augmentation methods. With the advancement of LLM and its nature of a multi-task learner~\cite{radford2019language}, we envision that a series of tasks in NLP can be enhanced or even replaced in a similar fashion.

Although AugGPT has shown promising results in data augmentation, it has certain limitations. For example, when recognizing and augmenting medical texts, AugGPT may produce incorrect augmentation results due to the lack of domain knowledge of ChatGPT. In future works, we will investigate adapting the general-domain LLMs, such as ChatGPT, to domain-specific data, such as medical texts, via model fine-tuning, in-context learning (prompt engineering), knowledge distillation, style transfer, etc.

AugGPT has demonstrated that the augmentation results can effectively improve the performance of the downstream classification task. A promising direction for future research is to investigate AugGPT against a wider range of downstream tasks. For example, given the strong ability of ChatGPT to extract key points and understand sentences, it can be utilized in tasks such as text summarization. Specifically, ChatGPT might be valuable for domain-specific science paper summarization ~\cite{9638337} and clinical report summarization ~\cite{CAI2022103999}. Publicly available domain-specific science paper summarization datasets and clinical report datasets are rare and often provided at small scales due to privacy concerns and the need for expert knowledge to generate annotated summaries. However, ChatGPT could address this challenge by generating diverse augmented summarization samples in different representation styles. The data generated from ChatGPT are typically concise, which can be valuable for further enhancing the generalization capabilities of the trained model.

The dramatic rise of generative image models such as DALLE2 ~\cite{ramesh2022hierarchical} and Stable Diffusion ~\cite{rombach2022high} provides opportunities for applying AugGPT to few-shot learning tasks in computer vision. For example, accurate language descriptions may be used to guide the generative model to generate images from text or to generate new images based on existing images as a data augmentation method for few-shot learning tasks, especially when combined with efficient fine-tuning methods ~\cite{hu2021lora,ruiz2022dreambooth} such as LoRA for Stable Diffusion. Thus, prior knowledge from a large language model can facilitate faster domain adaptation and better few-shot learning of generative models in computer vision.

Recent research shows that large language models (LLMs), such as GPT-3 and ChatGPT, are capable of solving Theory of Mind (ToM) tasks, which were previously thought to be unique to humans \cite{kosinski2023theory}. While the ToM-like capabilities of LLMs may be an unintended byproduct of improved performance, the underlying connection between cognitive science and the human brain is an area ripe for exploration. Advancements in cognitive and brain science can also be used to inspire and optimize the design of LLMs. For example, it has been suggested that the activation patterns of the neurons in the BERT model and those in the human brain networks may share similarities and could be coupled together\cite{liu2023coupling}. This presents a promising new direction for developing LLMs by utilizing prior knowledge from brain science. As researchers continue to investigate the connections between LLMs and the human brain, we may discover new means to enhance the performance and capabilities of AI systems, leading to exciting breakthroughs in the field.


%





\ifCLASSOPTIONcaptionsoff
  \newpage
\fi


\bibliography{ref}
\bibliographystyle{IEEEtran}
\end{document}


%
\title{AugGPT: Leveraging ChatGPT for Text Data Augmentation \\Supplementary Material}

\author{Haixing~Dai{$\ast$},
        Zhengliang~Liu{$\ast$},
        Wenxiong~Liao{$\ast$},
        Xiaoke~Huang,
        Yihan~Cao,
        Zihao~Wu,
        Lin~Zhao,
        Shaochen~Xu,
        Wei~Liu,
        Ninghao~Liu,
        Sheng~Li,
        Dajiang~Zhu,
        Hongmin~Cai, 
        Lichao~Sun,
        Quanzheng~Li,
        Dinggang~Shen,
        Tianming~Liu,
        and~Xiang~Li 
        
} 


\maketitle

\IEEEdisplaynontitleabstractindextext

%
\IEEEpeerreviewmaketitle

\begin{table*}[t]
\centering
\caption{ Generated Samples From ChatGPT on PubMed20K}
\label{table1}
{
\begin{tabular}{lll}
\hline
Original Text             & Altered Text                \\ \hline
\multirow{6}{*}{The follow-up rate after 5 years was 85\%.}                                                                & 1. The study showed a follow-up rate of 85\% after 5 years.                 \\
& 2. The 5-year follow-up rate was found to be significantly high at 85\%.    \\
& 3. The participants had an impressive follow-up rate of 85\% after 5 years. \\
& 4. The data indicated a follow-up rate of 85\% 5 years post-study.          \\
& 5. The 5-year follow-up rate was recorded as 85\%, according to the findings. \\
& 6. The results of the study showed a follow-up rate of 85\% after a 5-year period.  \\ \hline
\multirow{5}{*}{\begin{tabular}[c]{@{}l@{}}This means that the maturation of calluses \\ was 27\% faster in the LIPUS group.\end{tabular}}  & 1. The LIPUS group showed a 27\% acceleration in callus maturation.         \\
& \begin{tabular}[c]{@{}l@{}}2. The results revealed that the callus maturation process was 27\% faster in the LIPUS group.\end{tabular}  \\
& \begin{tabular}[c]{@{}l@{}}3. The study found that the callus maturation rate in the LIPUS group was 27\% quicker.\end{tabular}      \\
& \begin{tabular}[c]{@{}l@{}}4. Callus maturation was 27\% more rapid in the LIPUS group compared to others.\end{tabular} \\
& \begin{tabular}[c]{@{}l@{}}5. The callus maturation process in the LIPUS group was 27\% more efficient, according to the \\results.\end{tabular} \\ \hline
\multirow{6}{*}{\begin{tabular}[c]{@{}l@{}}Test Registration: IRCT.ir \\ IRCT2012071010230N1.\end{tabular}}                & \begin{tabular}[c]{@{}l@{}}1. The trial was registered with the IRCT under the identifier IRCT2012071010230N1.\end{tabular}                       \\
& \begin{tabular}[c]{@{}l@{}}2. The study was officially registered with IRCT.ir, with the registration number \\ IRCT2012071010230N1.\end{tabular} \\
& \begin{tabular}[c]{@{}l@{}}3. The trial was documented with IRCT.ir with the identification code IRCT2012071010230N1.\end{tabular}              \\
& \begin{tabular}[c]{@{}l@{}}4. IRCT.ir served as the official registrar for the trial, with the registration number \\IRCT2012071010230N1.\end{tabular}   \\
& \begin{tabular}[c]{@{}l@{}}5. The study was recorded with IRCT under the registration number IRCT2012071010230N1.\end{tabular}  \\
& 6. IRCT.ir recorded the trial with the identifier IRCT2012071010230N1.    \\ \hline
\multirow{6}{*}{\begin{tabular}[c]{@{}l@{}}While behavioral and technological \\ interventions can lead to some modest \\ improvements in glycemic control,  \\ these interventions have not performed \\ much better than conventional prevention \\ in achieving glycemic control.\end{tabular}} & \begin{tabular}[c]{@{}l@{}}1. The study found that although behavioral and technological interventions led to some \\ slight improvements in glycemic control, they were not significantly more effective \\ than typical care.\end{tabular}      \\
& \begin{tabular}[c]{@{}l@{}}2. Despite the modest improvement in glycemic control through behavioral and \\ technological interventions, they did not perform better than the standard care.\end{tabular}                                         \\
& \begin{tabular}[c]{@{}l@{}}3. The results showed that while behavioral and technological interventions resulted \\ in some minimal gains in glycemic control, they did not surpass the usual care in \\ achieving glycemic control.\end{tabular} \\
& \begin{tabular}[c]{@{}l@{}}4. Although behavioral and technological interventions showed some improvement \\ in glycemic control, they were not found to be significantly superior to the usual care.\end{tabular}                               \\
& \begin{tabular}[c]{@{}l@{}}5. The study showed that the usual care was not outperformed by behavioral and \\ technological interventions in terms of achieving glycemic control, despite some \\ small improvements.\end{tabular}                \\

 \hline
\end{tabular}
}
\end{table*}

\begin{table*}[]
\centering
\caption{ Generated Samples From ContextualWordAugUsingBert(insert\textbf{}) on PubMed20K}
\begin{tabular}{ll}
\hline
Original Text& Altered Text  \\ \hline
\multirow{6}{*}{The follow-up rate after 5 years was 85\%.}  & 1. follow - up rate even after compulsory 5 ¹⁄₂ years school was 85 \%.   \\
 & 2. their follow - off up rate repeated after following 5 years was 85 \%. \\
 & 3. her follow - was up rate after 5 years old was approximately 85 \%.\\
 & 4. follow - of up rate after just 5 years old was almost 85 \%.   \\
 & 5. his follow - follow up rate after 5 plus years was below 85 \%.\\
 & 6. follow - up pass rate only after 5 pre years pass was 85 \%.   \\ \hline
\multirow{6}{*}{\begin{tabular}[c]{@{}l@{}}This means that the maturation of calluses\\ was 27\% faster in the LIPUS group.\end{tabular}}& \begin{tabular}[c]{@{}l@{}}1. this means considering that each callus maturation was 27 \% the \\ faster in dividing the core lipus stem group.\end{tabular}  \\
 & \begin{tabular}[c]{@{}l@{}}2. this means that total callus mediated maturation progression was \\ often 27 \% faster in all the lipus member group.\end{tabular}  \\
 & \begin{tabular}[c]{@{}l@{}}3. all this means that callus male maturation rate was 27 \% faster \\ even in the original lipus study group.\end{tabular}\\
 & \begin{tabular}[c]{@{}l@{}}4. surprisingly this correlation means that human callus maturation\\  rates was almost 27 \% faster even in the lipus group.\end{tabular} \\
 & \begin{tabular}[c]{@{}l@{}}5. this evidence means that in callus group maturation was always \\ 27 \% or faster in entering the lipus group.\end{tabular} \\
 & \begin{tabular}[c]{@{}l@{}}6. this means rather that wild callus maturation was 27 \% or faster \\ found in the typical lipus species group.\end{tabular} \\ \hline
\multirow{6}{*}{\begin{tabular}[c]{@{}l@{}}While behavioral and technological\\ interventions can lead to some modest\\ improvements in glycemic control,\\ these interventions have not performed\\ much better than conventional prevention\\ in achieving glycemic control.\end{tabular}} & \begin{tabular}[c]{@{}l@{}}1. although some behavioral and improved technological interventions \\ can result partly in some comparatively modest improvements produced \\ in glycemic control, and these latter interventions did not fare around \\ significantly better than usual care in children achieving glycemic quality \\ control.\end{tabular}\\
 & \begin{tabular}[c]{@{}l@{}}2. but although behavioral and physical technological interventions can result \\ in in quite some modest outcomes improvements shown in glycemic control, \\ these interventions did apparently not fare more significantly at better than \\ for usual care in achieving glycemic control.\end{tabular}  \\
 & \begin{tabular}[c]{@{}l@{}}3. so although most behavioral and technological interventions can naturally \\ result in only some modest efficiency improvements dramatically in glycemic \\ control, these behavioral interventions sometimes did not generally fare \\ significantly better than usual care procedures in achieving glycemic control.\end{tabular} \\
 & \begin{tabular}[c]{@{}l@{}}4. and although mainstream behavioral and even technological therapy interventions \\ can theoretically result in some comparatively modest improvements in glycemic \\ control, these latter interventions did not fare any significantly better than most \\ usual care in doctors achieving glycemic control.\end{tabular}  \\
 & \begin{tabular}[c]{@{}l@{}}5. although conventional behavioral interventions and technological interventions \\ usually can result in doing some modest improvements still in glycemic control, \\ however these interventions typically did not fare significantly better results than \\ usual care systems in achieving glycemic lateral control.\end{tabular} \\
 & \begin{tabular}[c]{@{}l@{}}6. although behavioral adaptive and technological interventions themselves can \\ theoretically result in some modest potential improvements compared in glycemic \\ control, these new interventions did sometimes not ever fare significantly better \\ than usual health care in achieving glycemic damage control.\end{tabular}\\ \hline
\multirow{6}{*}{\begin{tabular}[c]{@{}l@{}}Test Registration: IRCT.ir\\ IRCT2012071010230N1.\end{tabular}}   & 1. clinical trial version registration : from irct. gov ir and irct2012071010230n1.   \\
 & 2. intermediate trial method registration : either irct. t ir or irct2012071010230n1. \\
 & 3. limited trial trial registration : standard irct. ru ir and irct2012071010230n1.   \\
 & 4. controlled trial model registration : both irct. br ir and irct2012071010230n1.\\
 & 5. human trial trials registration : in irct. ic ir code irct2012071010230n1. \\
 & 6. pre trial case registration : sri irct. ni ir code irct2012071010230n1.\\ \hline
\end{tabular}
\end{table*}

\begin{table*}[]
\centering
\caption{ Generated Samples From WordNetSynomymAug on PubMed20K}
\begin{tabular}{ll}
\hline
Original Text  & Altered Text      \\ \hline
The follow-up rate after 5 years was 85\%    & \begin{tabular}[c]{@{}l@{}}1. Follow - up pace after 5 twelvemonth be eighty five\%.\\ 2. Follow - up pace after 5 years be lxxxv\%.\\ 3. Follow - astir rate after fin year was 85\%.\\ 4. Follow - up charge per unit after 5 geezerhood was 85\%.\\ 5. Follow - upwards rate after five years exist 85\%.\\ 6. Follow - up rate after quintet year was lxxxv\%.\end{tabular}     \\ \hline
\begin{tabular}[c]{@{}l@{}}This means that the maturation of calluses\\ was 27\% faster in the LIPUS group.\end{tabular}                                                                                      & \begin{tabular}[c]{@{}l@{}}1. This means that callus maturation be twenty seven\% quicker in the LIPUS grouping.\\ 2. This means that callus maturation was xxvii\% quicker in the LIPUS group.\\ 3. This means that callus maturation was 27\% quicker in the LIPUS chemical group.\\ 4. This mean that callus development was twenty seven\% faster in the LIPUS chemical group.\\ 5. This mean that callus maturation be twenty seven\% faster in the LIPUS chemical group.\\ 6. This think that callus maturation be 27\% quicker in the LIPUS mathematical group.\end{tabular}                                                                                                                                                                                                                                                                                                                                                                                                                                                                                                                                                                                                                                                                                                                                                                                                                                   \\ \hline
\begin{tabular}[c]{@{}l@{}}While behavioral and technological\\ interventions can lead to some modest\\ improvements in glycemic control,\\ these interventions have not performed\\ much better than conventional prevention\\ in achieving glycemic control.\end{tabular}                                                                                                                                                                                  & \begin{tabular}[c]{@{}l@{}}1. Although behavioral and technological interventions give the axe result in some small \\ improvements in glycemic ascendence, these intervention did not do significantly good \\than  common care in achieving glycemic control."\\
2. Although behavioral and technical interposition can result in some pocket size \\improvements in glycemic ascendance, these interventions did non fare significantly \\better than usual fear in accomplish glycemic control."\\ 
3. Although behavioural and technical intervention can leave in some modest \\improvement in glycemic control, these interventions execute not fare importantly \\expert than usual care in achieving glycemic mastery."\\ 
4. Although behavioural and technical interventions rear end result in some modest \\improvements in glycemic control, these interventions get along not do importantly \\good than usual care in achieving glycemic control."\\ 
5. Although behavioural and technological treatment backside result in some modest \\improvements in glycemic control, these interventions did non fare importantly better \\than usual maintenance in achieve glycemic control."\\ 
6. Although behavioural and technological interventions can result in some small \\betterment in glycemic control, these interventions do non fare importantly better than \\usual care in achieve glycemic control."\end{tabular} \\ \hline
\begin{tabular}[c]{@{}l@{}}Test Registration: IRCT.ir\\ IRCT2012071010230N1.\end{tabular} 
 & \begin{tabular}[c]{@{}l@{}}1. Trial Registration: IRCT. inland revenue IRCT2012071010230N1.\\ 2. Trial Registration: IRCT. atomic number 77 IRCT2012071010230N1.\\ 3. Trial Readjustment: IRCT. inland revenue IRCT2012071010230N1.\\ 4. Trial Enrollment: IRCT. ir IRCT2012071010230N1.\\ 5. Trial Readjustment: IRCT. ir IRCT2012071010230N1.\\ 6. Trial Registration: IRCT. ir IRCT2012071010230N1.\end{tabular} \\ \hline
\end{tabular}
\end{table*}

\begin{table*}[]
\centering
\caption{ Generated Samples From ChatGPT on Symtoms}
\begin{tabular}{ll}
\hline
Original Text                                   & Altered Text                                                                                                                                                                                                                                                                                                                                                                                                                                                                                                                                                                                                                      \\ \hline
I fell a stiffness in my elbows and shoulders & \begin{tabular}[c]{@{}l@{}}1. The speaker is experiencing stiffness in their elbows and shoulders.\\ 2. The speaker mentions feeling a stiffness in their elbows and shoulders.\\ 3. The speaker reports a stiffness in their elbows and shoulders.\\ 4. The speaker has noticed a stiffness in their elbows and shoulders.\\ 5. The speaker feels uncomfortable due to the stiffness in their elbows and shoulders.\\ 6. The speaker describes a stiffness in their elbows and shoulders that is causing discomfort.\end{tabular}                                                                                    \\ \hline
I have pain in the elbow joint                & \begin{tabular}[c]{@{}l@{}}1. The speaker is experiencing pain in their elbow joint.\\ 2. The speaker mentions having pain in their elbow joint.\\ 3. The speaker reports pain in their elbow joint.\\ 4. The speaker has noticed pain in their elbow joint.\\ 5. The speaker feels discomfort due to the pain in their elbow joint.\\ 6. The speaker describes the pain in their elbow joint that is causing discomfort.\end{tabular}                                                                                                                                                                                \\ \hline
My son has a lot of acne.                     & \begin{tabular}[c]{@{}l@{}}1. The speaker's son has a high number of acne breakouts.\\ 2. The speaker mentions that their son has many acne pimples.\\ 3. The speaker's son is struggling with a lot of acne on their skin.\\ 4. The speaker has noticed a significant amount of acne on their son's face.\\ 5. The speaker's son's skin is affected by a large number of acne pimples.\\ 6. The speaker describes their son's skin as having a lot of acne.\end{tabular}                                                                                                                                             \\ \hline
I can't have a girlfriend because of my acne. & \begin{tabular}[c]{@{}l@{}}1. The speaker believes that their acne is the reason they cannot find a girlfriend.\\ 2. The speaker feels that their acne is preventing them from getting a girlfriend.\\ 3. The speaker mentions that their inability to find a girlfriend is due to their acne.\\ 4. The speaker thinks that their acne is a hindrance in their quest for a girlfriend.\\ 5. The speaker is having trouble finding a girlfriend because of their acne-prone skin.\\ 6. The speaker believes that their acne is impacting their dating life and preventing \\them from finding a girlfriend.\end{tabular} \\ \hline
I feel fluid when I cough.                    & \begin{tabular}[c]{@{}l@{}}1. The speaker is experiencing the sensation of fluid rising when they cough.\\ 2. The speaker feels like something is coming up when they cough.\\ 3. The speaker coughs and feels like fluid is trying to escape.\\ 4. The speaker has a fluid sensation in their throat when they cough.\\ 5. When coughing, the speaker feels as if fluid is attempting to rise.\\ 6. The speaker is experiencing an upward flow of fluid when coughing.\end{tabular}                                                                                                                                  \\ \hline
I can't breathe                               & \begin{tabular}[c]{@{}l@{}}1. The speaker is experiencing difficulty breathing.\\ 2. The speaker is finding it hard to inhale air.\\ 3. The speaker is gasping for air.\\ 4. Breathing is a struggle for the speaker.\\ 5. The speaker is experiencing respiratory distress.\\ 6. The speaker feels suffocated and can't breathe properly.\end{tabular}                                                                                                                                                                                                                                                               \\ \hline
\end{tabular}
\end{table*}

\begin{table*}[]
\centering
\caption{ Generated Samples From ContextualWordAugUsingBert(insert\textbf{}) on Symtoms}
\begin{tabular}{ll}
\hline
Original Text                                   & Altered Text                                                                                                                                                                                                                                                                                                                                                                                                                                                                                           \\ \hline
I fell a stiffness in my elbows and shoulders & \begin{tabular}[c]{@{}l@{}}1. i fell a delicious stiffness in my skinny elbows and back shoulders\\ 2. eventually i fell from a stiffness in my elbows and i shoulders\\ 3. i fell in a growing stiffness born in my elbows and shoulders\\ 4. everywhere i fell a stiff stiffness in my elbows neck and shoulders\\ 5. i fell a white stiffness growing in my elbows together and shoulders\\ 6. i just fell a stiffness up in my elbows and lower shoulders\end{tabular}                 \\ \hline
I have pain in the elbow joint                & \begin{tabular}[c]{@{}l@{}}1. except i do have a pain in my left elbow joint\\ 2. i have felt a pain beginning in my elbow or joint\\ 3. i have a great pain in all my bad elbow joint\\ 4. i have given a pain in putting my elbow ball joint\\ 5. thankfully i have also a faint pain in my elbow joint\\ 6. today i have a pain bite in my middle elbow joint\end{tabular}                                                                                                              \\ \hline
My son has a lot of acne.                     & \begin{tabular}[c]{@{}l@{}}1. my eldest son has saved a lot of local acne.\\ 2. but my son has a damn lot consisting of acne.\\ 3. my older son matt has a lot kind of acne.\\ 4. my beloved son has a lot kind of new acne.\\ 5. apparently my best son has sustained a lot of acne.\\ 6. little my dear son has a nice lot of acne.\end{tabular}                                                                                                                                         \\ \hline
I can't have a girlfriend because of my acne. & \begin{tabular}[c]{@{}l@{}}1. and i really can't go get a girlfriend because of my bad acne.\\ 2. i can'from t get a girlfriend anymore because of wearing my own acne.\\ 3. like i can'bal t get a healthy girlfriend because of having my acne.\\ 4. sometimes i still can'of t get a girlfriend because of my chronic acne.\\ 5. i also can't really get a serious girlfriend because of my first acne.\\ 6. maybe i can'′ t get a girlfriend here because of my own acne.\end{tabular} \\ \hline
I feel fluid when I cough.                    & \begin{tabular}[c]{@{}l@{}}1.suddenly i feel some fluid trying to again come back up when i cough.\\ 2.i even feel fluid trying constantly to clean come him up when i cough.\\ 3.i still feel the fluid fluids trying to come up when suddenly i cough.\\ 4.i feel his fluid trying to come right up breathe when finally i cough.\\ 5.i often feel fluid trying to still come clean up right when i cough.\\ 6.i feel fluid trying me to come waking up when i i say cough.\end{tabular} \\ \hline
I can't breathe                               & \begin{tabular}[c]{@{}l@{}}1.i also cant even breathe\\ 2.i just cant freaking breathe\\ 3.yet i still cant breathe\\ 4.now i almost cant breathe\\ 5.how i cant fully breathe\\ 6.pity i cant than breathe\end{tabular}                                                                                                                                                                                                                                                                   \\ \hline
\end{tabular}
\end{table*}

\begin{table*}[]
\centering
\caption{ Generated Samples From WordNetSynomymAug on Symtoms}
\begin{tabular}{ll}
\hline
Original Text                                   & Altered Text                                                                                                                                                                                                                                                                                                                                                                                                 \\ \hline
I fell a stiffness in my elbows and shoulders & \begin{tabular}[c]{@{}l@{}}1. I devolve a rigorousness in my elbows and shoulders\\ 2. Iodin fell a stiffness in my elbow and shoulders\\ 3. I fell a stiffness in my elbows and shoulder\\ 4. Iodine shine a stiffness in my human elbow and shoulders\\ 5. I fell a stiffness in my cubitus and shoulders\\ 6. I fell a stiffness in my elbow and shoulders\end{tabular}                       \\ \hline
I have pain in the elbow joint                & \begin{tabular}[c]{@{}l@{}}1. i have a pain in my articulatio cubiti joint\\ 2. i sustain a pain in my elbow roast\\ 3. i hold a botheration in my elbow reefer\\ 4. i hold a pain in my human elbow joint\\ 5. iodine have a pain in my elbow stick\\ 6. one sustain a pain in the neck in my elbow joint\end{tabular}                                                                          \\ \hline
My son has a lot of acne.                     & \begin{tabular}[c]{@{}l@{}}1. My son throw a lot of acne.\\ 2. My word has a batch of acne.\\ 3. My boy have a lot of acne.\\ 4. My son have a lot of acne.\\ 5. My son have a destiny of acne.\\ 6. My word has a lot of acne.\end{tabular}                                                                                                                                                     \\ \hline
I can't have a girlfriend because of my acne. & \begin{tabular}[c]{@{}l@{}}1. Atomic number 53 lav ' t get a girlfriend because of my acne.\\ 2. I potty ' t get a lady friend because of my acne.\\ 3. Ace fire ' t get a lady friend because of my acne.\\ 4. One can ' t get a lady friend because of my acne.\\ 5. Iodin butt ' t get a girlfriend because of my acne.\\ 6. I bum ' t generate a girl because of my acne.\end{tabular}       \\ \hline
I feel fluid when I cough.                    & \begin{tabular}[c]{@{}l@{}}1. I feel fluid render to arrive up when I cough.\\ 2. I feel fluid try to come up when 1 cough.\\ 3. I feel fluid trying to fall upwards when I cough.\\ 4. Iodin feel fluid trying to amount up when I cough.\\ 5. I palpate fluid trying to get along upwardly when Ace cough.\\ 6. I sense fluid trying to come upwards when Atomic number 53 cough.\end{tabular} \\ \hline
I can't breathe                               & \begin{tabular}[c]{@{}l@{}}1. unity cant breathe\\ 2. i cant breathe\\ 3. atomic number 53 cant breathe\\ 4. i cant breathe\\ 5. i pitch breathe\\ 6. i pitch breathe\end{tabular}                                                                                                                                                                                                               \\ \hline
\end{tabular}
\end{table*}